\documentclass{article}
\usepackage{spconf,amsmath,graphicx}
\usepackage{multirow,graphicx}
\usepackage{amsmath}
\usepackage{color}
\usepackage{subfigure}
\usepackage{amssymb}
\usepackage{array}%
\usepackage{makecell}
\usepackage{algorithm}  
\usepackage{algorithmicx}  
\usepackage{algpseudocode}
\usepackage{bbding}
\usepackage{amssymb}
\usepackage{amsmath,graphicx,multirow,bm}
\usepackage[colorlinks,linkcolor=black]{hyperref}

\title{PAN-SHARPENING VIA HIGH-PASS MODIFICATION CONVOLUTIONAL NEURAL NETWORK}
\name{Jiaming Wang $^{1}$, Zhenfeng Shao $^{1}$\thanks{This work was supported in part by the National Key R\&D Program of China under Grant 2018YFB0505401; in part by the National Natural Science Foundation of China under Grants 41890820, 41771452, 41771454, and 62072350; in part by the Key R\&D Program of Yunnan Province in China under Grant 2018IB023.}, Xiao Huang $^{2}$, Tao Lu $^{3}$, Ruiqian Zhang $^{1}$, Jiayi Ma $^{4}$}
\address{$^{1}$ State Key Laboratory for Information Engineering in Surveying, Mapping and Remote Sensing, \\Wuhan University\\
	$^{2}$ Department of Geosciences, University of Arkansas\\
$^{3}$Hubei Key Laboratory of Intelligent Robot, School of Computer Science and Engineering, \\Wuhan Institute of Technology,\\
$^{4}$ the Electronic Information School, Wuhan University\\}
\begin{document}
%
\maketitle
\begin{abstract}
Most existing deep learning-based pan-sharpening methods have several widely recognized issues, such as spectral distortion and insufficient spatial texture enhancement, we propose a novel pan-sharpening convolutional neural network based on a high-pass modification block. Different from existing methods, the proposed block is designed to learn the high-pass information, leading to enhance spatial information in each band of the multi-spectral-resolution images. To facilitate the generation of visually appealing pan-sharpened images, we propose a perceptual loss function and further optimize the model based on high-level features in the near-infrared space. Experiments demonstrate the superior performance of the proposed method compared to the state-of-the-art pan-sharpening methods, both quantitatively and qualitatively. The proposed model is open-sourced at \url{https://github.com/jiaming-wang/HMB}.

\end{abstract}
\vspace{-0.4cm}
\section{Introduction}
\label{sec:intro}

With the rapid development of satellite sensors, remote sensing images have become widely used in various academic fields and in both civil and military domains. However, as recognized by many, remote sensing sensors own the intrinsic trade-off between the spatial resolution and spectral resolution \cite{shen2016integrated}. As a result, for a certain remote sensing sensor, the captured gray panchromatic (PAN) images are generally with a finer resolution than multispectral (MS) images. Typically, the goal of pan-sharpening is to integrate the texture details of the high-resolution (HR) PAN images and the spectral information low-resolution (LR) MS images for the purpose of generating HR-MS images, thus breaking the technological limits. Therefore, pan-sharpening tasks can be regarded as a crucial pre-processing step for numerous remote sensing data applications \cite{cai2020super}.

Over the past few decades, many different pan-sharpening algorithms have been proposed. Traditional pan-sharpening methods can be divided into three main categories: component substitution (CS)-, multi-resolution analysis (MRA)-, and variational optimization (VO)-based methods \cite{meng2019review}. CS-based methods aim to separate the spatial and spectral information of the MS images in a suitable space and further fuse them with the corresponding PAN images. The representative CS-based methods include the Intensity-Hue-Saturation (IHS) methods, the principal component analysis (PCA) methods, and the Gram-Schmidt (GS) methods \cite{laben2000process}. In comparison, MRA-based methods decompose MS and PAN images into the multi-scale space via the utilization of the Laplacian pyramid transform and the Wavelet transform. The decomposed PAN images are injected into the corresponding MS images for information fusion \cite{aiazzi2002context}. VO-based methods regard the pan-sharpening tasks as an ill-posed problem \cite{aly2014regularized} with the goal of loss minimization. Li \emph{et al.} \cite{li2010new} first proposed a sparse-representation pan-sharpening method that reconstructed the HR-MS images using similar image patches of the dictionary in the PAN space. Further, Cheng \emph{et al.} \cite{cheng2013sparse} adopted a two-step learning strategy with improved representation ability. Following the above pioneering works, various regularization terms were introduced to the representation coefficient optimization. However, the performance of these traditional methods is limited due to the models’ shallow nonlinear expression ability.


Recently, several convolutional neural networks-based (CNN) pan-sharpening algorithms have been proposed, benefiting from the powerful feature extraction capability of CNN and the development of deep learning-based single-image super-resolution techniques that recover the missing high-frequency information from the LR image. Inspired by the super-resolution CNN (SRCNN) \cite{dong2015image}, PNN \cite{masi2016pansharpening} learned the nonlinear mapping function from the concatenation of the LR-MS and PAN to the HR-MS image. Scarpa \emph{et al.} \cite{scarpa2018target} proposed a target-adaptive CNN-based framework that learned the residual image with improved performance over image fusion tasks. Multi-scale and multi-depth CNN (MSDCNN) \cite{yuan2018multiscale} was proposed to extract multi-scale features. Han \emph{et al.} \cite{xu2020sdpnet} proposed a surface- and deep-level constraint-based pan-sharpening network. Cai \emph{et al.} \cite{cai2020super} proposed a super-resolution guided progressive pan-sharpening CNN, achieving better performance by combining multi-scale features.

Despite the first-rate results these CNN-based methods can obtain, some problems remain to be solved. As most of these methods are inspired by the concept of single-image super-resolution that generates an HR image from the corresponding LR one, they fail to make full use of the rich high-frequency spatial information from HR-PAN images. Some methods \cite{jiang2020differential,cai2020super} treat the PAN image as the missing high-frequency information of the MS image, nevertheless leading to spectral distortion \cite{jiang2020differential}. 
To address these problems, we proposed a novel pan-sharpening method based on a high-pass modification block, termed HMCNN. The main contributions of the proposed method are:
\begin{enumerate}
\item We design a high-pass modification block to enhance the spatial information in each band of the MS image. The corrected residual image can fully exploit the spatial prior of the PAN images, largely avoiding the problem of spectral distortion.
\item We use a perceptual loss to constrain the model in the near-infrared (NIR) band. These feature maps in the perceptual domain play an essential role in the detail enhancement of texture information.
\end{enumerate}
\vspace{-0.6cm}
\section{Method}\label{s2}
\vspace{-0.3cm}
\subsection{Problem Formulation}
\begin{figure}[h]
	\centering
	\includegraphics[height=4cm]{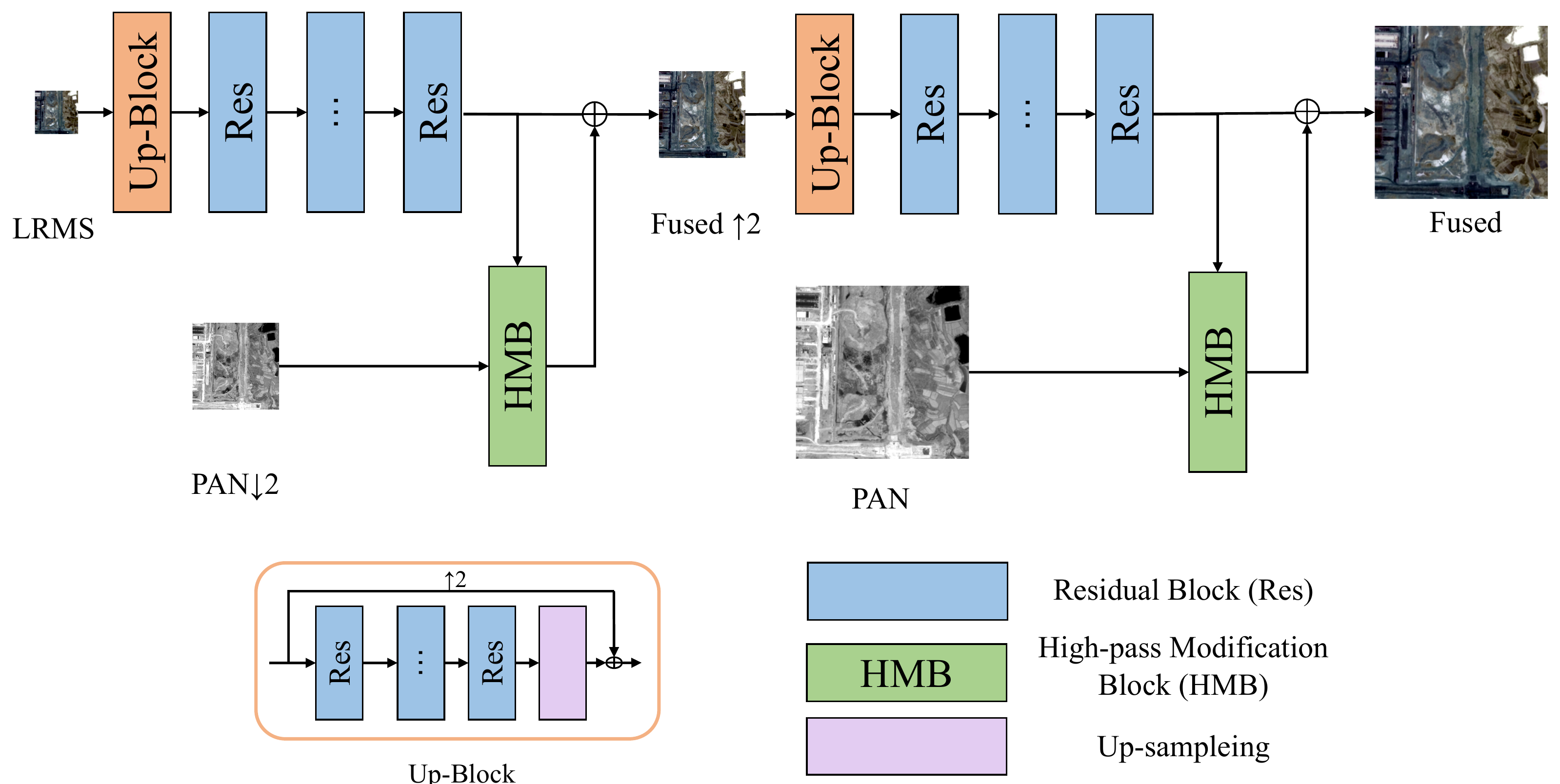}
	\caption{The overall structure of the proposed HMCNN. $\oplus$ denotes element-wise sum.}
	\label{network}
\end{figure}

Fig. \ref{network} presents the flowchart of the proposed HMCNN. We intent to super-resolve the input LR-MS with a multi-scale framework to alleviate the burden of computing resources. We denote the LR-MS image as $I_{LRMS}$ of size $w \times h  \times c$ and the corresponding HR-PAN image as $I_{PAN}$ of size $sw \times sh \times 1$, and the ground truth HR-MS $I_{HRMS}$ of size $sw \times sh  \times c$ , where $w$ and $h$ represent the width and height of the LR-MS image, respectively. $c$ is the number of image bands in the MS image, and $s$ denotes the spatial resolution scale between $I_{LRMS}$ and $I_{PAN}$. Our goal is to generate the HR-MS image $I_{Fused}$ from the input HR-PAN image $I_{PAN}$ and the LR-MS image $I_{LRMS}$ via an end-to-end solution,
\begin{equation}
\begin{aligned}
{I_{Fused \uparrow 2}} = {f_1}({I_{LRMS}}) + {f_{HMB}}({f_1}({I_{LRMS}}),{I_{PAN \downarrow 2}}),
\end{aligned}
\end{equation}
\begin{equation}
\begin{aligned}
{I_{Fused}} = {f_2}({I_{LRMS \uparrow 2}}) + {f_{HMB}}({f_2}({I_{LRMS \uparrow 2}}),{I_{PAN}}),
\end{aligned}
\end{equation}
where $H_{HMB}(.)$ refers to the proposed high-pass modification block (HMB). $f_{1}(.)$ and $f_{2}(.)$ denote the CNN-based feature extractor. $I_{PAN \downarrow 2}$ denotes the down-sampling version of $I_{PAN}$ with scale factor $\times$ 2 via a bicubic interpolation function. $I_{Fused}$ and ${I_{Fused \uparrow 2}}$ are the fused HR-MS image (scale factor $\times$ 4) and the fused HR-MS image (scale factor $\times$ 2), respectively.
\vspace{-0.6cm}
\subsection{The High-pass Modification Block (HMB)}
Pan-sharpening belongs to an ill-posed problem, which employs additional prior from PAN images for the reconstruction. Traditional approaches \cite{jiang2020differential,cai2020super} try to exploit the high-frequency information as a residual map. As for the MS image enhancement problem, it is crucial to effectively exploit the intrinsic relations between the PAN images and all bands of the MS image. Inspired by Zhang \emph{et al.} \cite{zhang2018image}, we introduce a spatial attention mechanism to adaptively rescale each pixel-wise feature, thus improving the representation ability of the spatial residual information .

\begin{figure}[h]
	\centering
	\includegraphics[height=4cm]{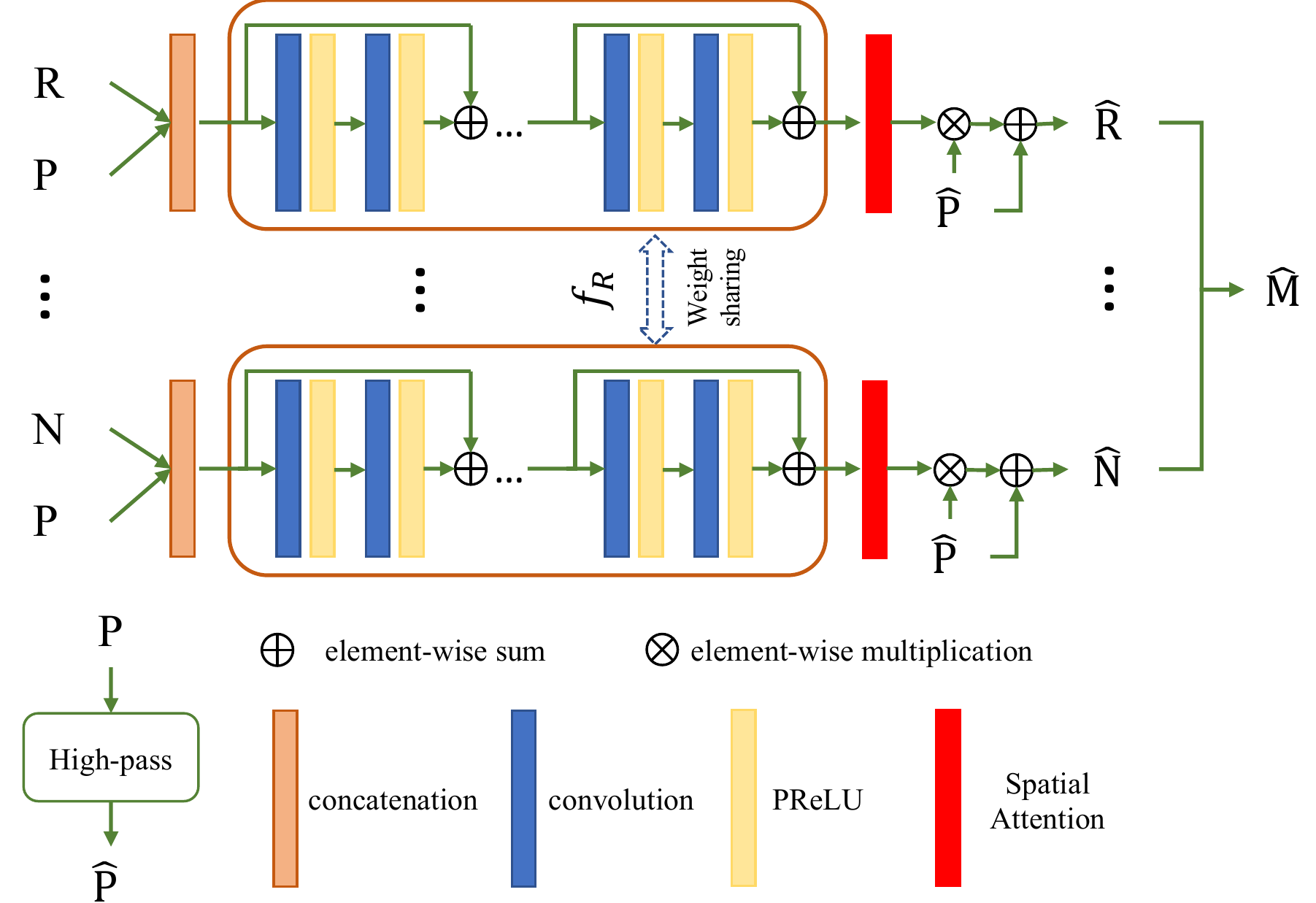}
	\caption{The architecture of high-pass modification block.}
	\label{rm}
\end{figure}

The proposed high-pass modification block (Fig. \ref{rm}), termed HMB, is designed to exploit the high-frequency prior. In particular, HMB cascade $n$ ($n$ = 11) residual blocks with $3 \times 3$ convolutions.
\begin{equation}
{F_{X,1}} = {f_R}(concat(X,P)),
\end{equation}
where $concat(.)$ refers to the concatenation function and $f_{R}(.)$ denotes the residul network. $X$ denotes the band in the MS image ($X \in \{ R,G,B,N\}$, N is NIR). We further feed ${F_{X,1}}$ into a spatial attention block,
\begin{equation}
\widehat P = P - D(P),
\end{equation}
\begin{equation}
\widehat X = ({f_{SA}}({F_{X,1}}) + 1)\widehat P,
\end{equation}
where $D(.)$ is the bicubic upsampling operation and $\widehat P$ refers to the high-pass version (residual information) of the PAN image. $f_{SA}(.)$ denotes the function of the spatial attention module. $\widehat X$ denotes the high-pass version of $X$ in the MS image. We obtain the output of the current HMB by
\begin{equation}
{\widehat M} = concat(\widehat R, \widehat G, \widehat B, \widehat N),
\end{equation}
where ${\widehat M}$ denotes the high-pass version of the MS image that is used for the reconstruction.

\vspace{-0.5cm}
\subsection{The Loss Function}
Existing pan-sharpening methods generally take pixel-wise error as the loss function for higher objective assessment results. Among them, mean square error (MSE) and mean absolute error (MAE) are the two most widely used loss functions. Specifically, the $\ell_2$ loss is defined as the MSE between the HR-MS and ground truth:
\begin{equation}
{\mathcal{L} _1}(\theta) = \frac{1}{M}\sum\limits_{m = 1}^N {{{\left\| {I_{HRMS}^m - {I_{Fused}^m}} \right\|}_2}},
\end{equation}
where $I_{Fused}^m$ is the $m$-th fused HR-MS. $M$ denotes the number of images in a training batch. $\theta$ denotes the parameters in HMCNN.

Although pixel-wise loss functions can achieve results with higher objective evaluations, they tend to generate over-smoothed images and often fail to capture perceptual differences (e.g., information regarding the textures and edges). Inspired by the VGG-based feature-wise loss function in SR tasks, we propose a novel perceptual loss (NIR-VGG) for texture enhancement, specifically targeting the NIR band. Different from the natural images scene, the spectral wavelength of the NIR band of MS images is larger than other bands, thus providing abundant texture information. Specifically, the NIR-VGG loss minimizes the feature-wise error in the NIR space,
\begin{equation} 
\begin{aligned}
{{\cal L}_{NN}}(\theta ) = & \frac{1}{N}\sum\limits_{n = 1}^N {\left\| \phi({NIR(I_{Fused}^n)} ) \right.} \\
&{{{\left. { - \phi(NIR({I_{HRMS}^n}))} \right\|}_2},}
\end{aligned}
\end{equation}
where $NIR(.)$ denotes the NIR band of an image and $\phi(.)$ is the output feature maps of the pre-trained VGG-19 network. ${\cal L}_{NN}$ indicates the loss bwtween HR-MS and fused images in the NIR band space. 
The final objective loss is simply computed by
\begin{equation}
{\mathcal{L}_{total}}(\theta) = \alpha {\mathcal{L}_{NN}}(\theta) + {\mathcal{L} _1}(\theta),
\end{equation}
where $\alpha$ is used to balance the contributions of different losses. In our experiments, we set $\alpha$ as a constant: $ \alpha  = 1e -3$.
\vspace{-0.6cm}
\section{Experiments}\label{s3}
\vspace{-0.4cm}
\subsection{Dataset and Implementation Details}
We conduct experiments on the WordView II (WV) dataset. The spatial resolution of PAN images in the dataset is 0.5 m with a spatial resolution ratio of 4 ($s=4$) between PAN and LR-MS images. We crop the HR-MS images (the original MS) into image patches of size $256 \times 256 \times 4$ pixels and the PAN images (the downsampled version of original PAN with the scale factor $\times 4$) into image patches of size $256 \times 256 \times 1$ pixels. Then, the downsampled LR-MS image patches are with a size of $64 \times 64 \times 4$ pixels.

The proposed model is trained with a batch size $12$ of $\ell _2$. Adam optimizer is used to optimize the proposed model with $\beta_1 =0.9$, $\beta_2 =0.999$, and $\varepsilon=1e-8$. The learning rate is initialized to $1e -4$, and decreases by a factor of 10 for every 1,000 epochs. The training stops at 2,000 epochs. We use PyTorch on a desktop with Ubuntu 18.04, CUDA 10.2, CUDNN 7.5, and Nvidia TAITAN RTX GPUs.

Eight widely used metrics of image quality assessment are employed to evaluate the performance, including erreur relative dimensionless global error in synthesis (ERGAS), root mean square error (RMSE), root mean absolute error (RMAE), spectral angle mapper (SAM), a universal image quality index (UIQI), the spectral distortion index $D_\lambda$, the spatial distortion index $D_S$, and the quality without reference (QNR). The best values for these metrics are 0, 0, 0, 0, 1, 0, 0, and 1, respectively. Among the above metrics, the last three metrics (i.e., $D_{\lambda}$, $D_{S}$, QNR) are no-reference image quality assessment. For RMAE, RMSE, and UIQI of the pan-sharpening images, we report their mean values of all bands.

\begin{table*}[htbp]
	\centering
	\normalsize
	\renewcommand\arraystretch{1}
	\caption{Comparison of pan-sharpening performance among the HMCNN and eight other state-of-the-art methods on the WV dataset. Best and second-best results are highlighted by \textcolor{red}{red} and \textcolor{blue}{blue}, respectively.}
	\begin{tabular}{c|cccccccc}
		\hline
		Methods & ERGAS $\downarrow$ & RMSE  $\downarrow$ & RMAE $\downarrow$  & SAM $\downarrow$ & UIQI $\uparrow$ & $D_{\lambda}$ $\downarrow$& $D_{S}$ $\downarrow$ & QNR $\uparrow$ \\
		\hline
		IHS \cite{carper1990use}   & 2.0743  & 4.7351  & 8.2615  & 2.6611  & 0.5540  & 0.0774  & 0.2059  & 0.7337  \\
		GS  \cite{laben2000process}    & 1.9827  & 4.7200  & 8.2133  & 2.5514  & 0.6070  & 0.0387  & 0.1536  & 0.8147  \\
		Brovey \cite{gillespie1987color} & 1.8615  & 4.4840  & 7.8146  & 2.2997  & 0.6017  & 0.0629  & 0.1706  & 0.7780  \\
		SFIM \cite{liu2000smoothing}   & 2.5265  & 5.6198  & 9.7548  & 2.5775  & 0.5897  & 0.0842  & 0.1407  & 0.7875  \\
		PNN \cite{masi2016pansharpening}   & 1.9399  & 4.4261  & 7.7268  & 2.5682  & 0.5875  & 0.0321  & 0.0790  & 0.8915  \\
		PANNet \cite{scarpa2018target}  & 1.3243  & 3.2157  & 5.6180  & 1.9270  & 0.6700  & 0.0396  & 0.1072  & 0.8575  \\
		MSDCNN \cite{yuan2018multiscale} & 1.3847  & 3.2541  & 5.6762  & 1.8307  & 0.6903  & 0.0407  & 0.0919  & 0.8713  \\
		SRPPNN \cite{cai2020super} & \textcolor{blue}{1.2061}  & \textcolor{blue}{2.8147}  & \textcolor{blue}{4.9074}  & \textcolor{blue}{1.5292}  & \textcolor{blue}{0.7198}  & \textcolor{blue}{0.0187}  & \textcolor{blue}{0.0720}  & \textcolor{blue}{0.9107}  \\
		HMCNN(our) &  \textcolor{red}{1.0033}  &  \textcolor{red}{2.6027}  &  \textcolor{red}{4.5377}  &  \textcolor{red}{1.4244}  &  \textcolor{red}{0.7513}  &  \textcolor{red}{0.0163}  &  \textcolor{red}{0.0698}  &  \textcolor{red}{0.9150}  \\
		\hline
	\end{tabular}%
	\label{twv}%
\end{table*}%

\vspace{-0.3cm}
\subsection{Comparison with State-of-The-Art Methods}

To verify the effectiveness of the HMCNN, eight state-of-the-art pan-sharpening algorithms are selected for the comparison: IHS \cite{carper1990use}, GS \cite{laben2000process}, Brovey \cite{gillespie1987color}, SFIM \cite{liu2000smoothing}, PNN \cite{masi2016pansharpening}, PANNet \cite{yang2017pannet}, MSDCNN \cite{yuan2018multiscale}, and SRPPNN \cite{cai2020super}. 

\begin{figure}[h]
	\centering
	\includegraphics[height=4cm]{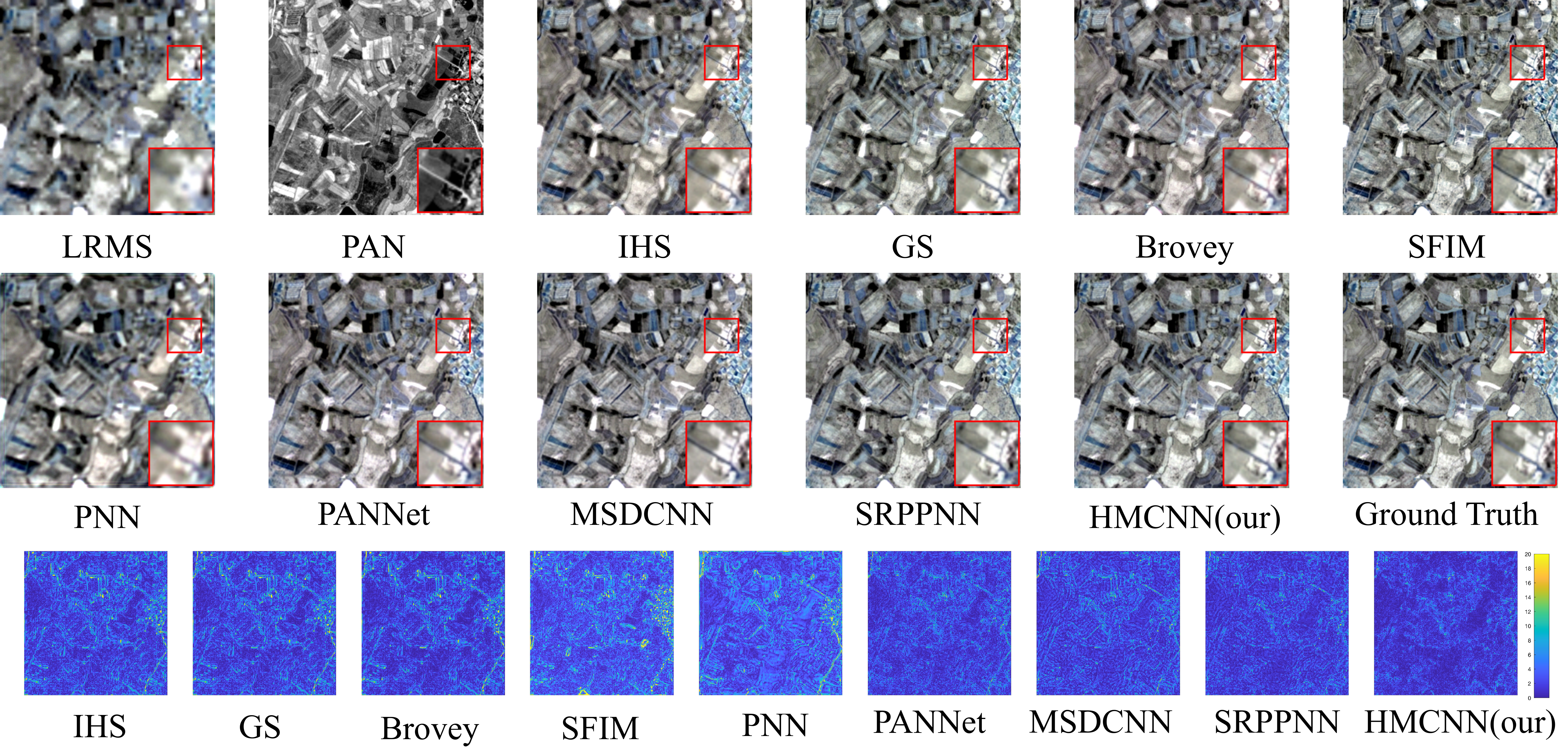}
	\caption{Qualitative comparison of HMCNN with eight comparative algorithms on a typical satellite image pair from the WV dataset. Images in the last row row represent the MSE between the pan-sharpened images and the ground truth.}
	\label{wv}
\end{figure}

In Table \ref{twv}, we report the quantitative evaluation results of the these architectures over the aforementioned evaluation metrics, where the number highlighted by red represents the best result in each evaluation metric and the blue represents the second best. We observe that HMCNN surpasses SRPPNN in all the indices. Quantitative evaluation (Fig. \ref{wv}) also proves the superiority of our method. Images in the last row of Fig. \ref{wv} are the MSE between the pan-sharpened results and the ground truth. By comparing pan-sharpened images with the ground truth, we observe that traditional algorithms perform poorly with noticeable artifacts. The pan-sharped images using HMCNN and SRPPNN are similar to the ground truth comparing with other deep learning based methods , evidenced by their error maps. In the zoomed area (highlighted using red rectangles in Fig. \ref{wv}), HMCNN successfully preserves the rich texture information of land and ditches. 

\begin{figure}[h]
	\centering
	\includegraphics[width=7cm]{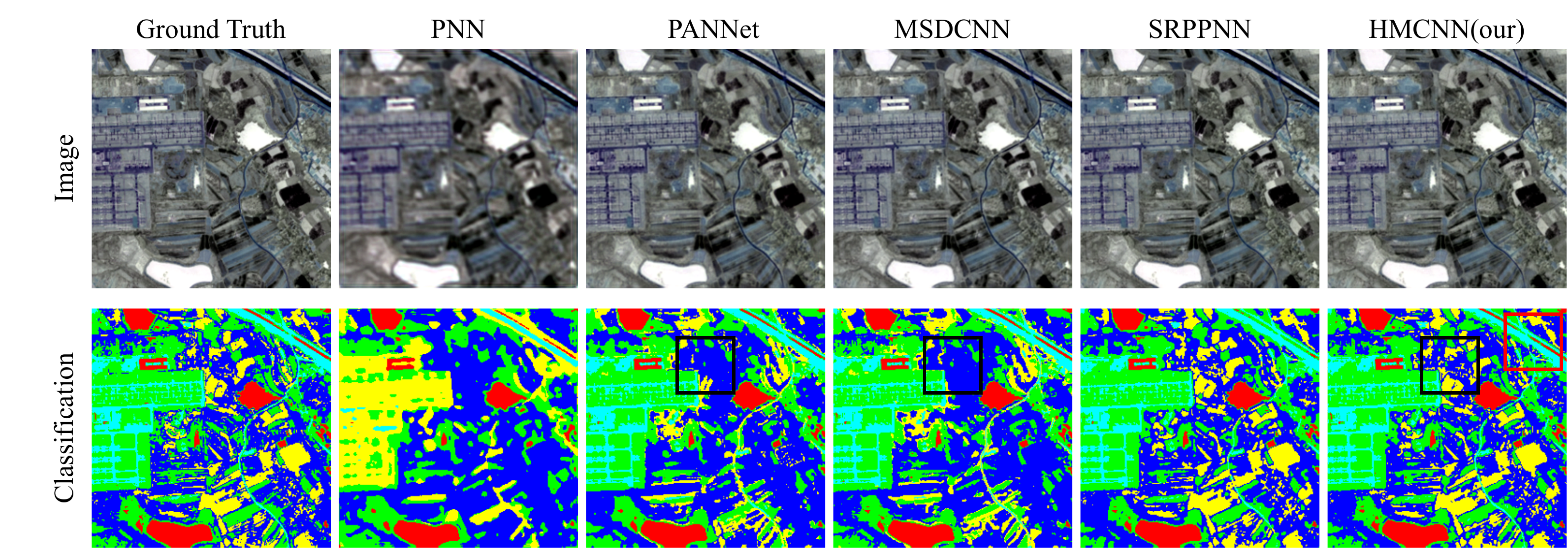}
	\caption{Pan-sharpened images (the first row) and the classification results using the ISODATA classification method (the second row). The boxes highlight the differences among various approaches and emphasize the superiority of HMCNN.}
	\label{classification}
\end{figure}

\vspace{-0.4cm}
We employ ISODATA, a classical unsupervised semantic segmentation using an iterative self-organizing techniques, to evaluate the results of different deep learning based pan-sharpening methods. The number of classifications in the ISODATA is set to five. We set the maximum iteration to five times. Fig. \ref{classification} shows the pan-sharpened (the first row) and classification results using ISODATA (the second row) with the boxes highlighting the differences among various approaches and emphasizing the superiority of HMCNN. We observe that PNN tends to generate blurred pan-sharpened images, thus preventing a fine classification result. The classification result of HMCNN includes considerably more accurate details and is visually the closest to the ground truth.

\vspace{-0.3cm}
\subsection{Running Time Analysis}
\begin{figure}[h]
	\centering
	\includegraphics[width=3.7cm]{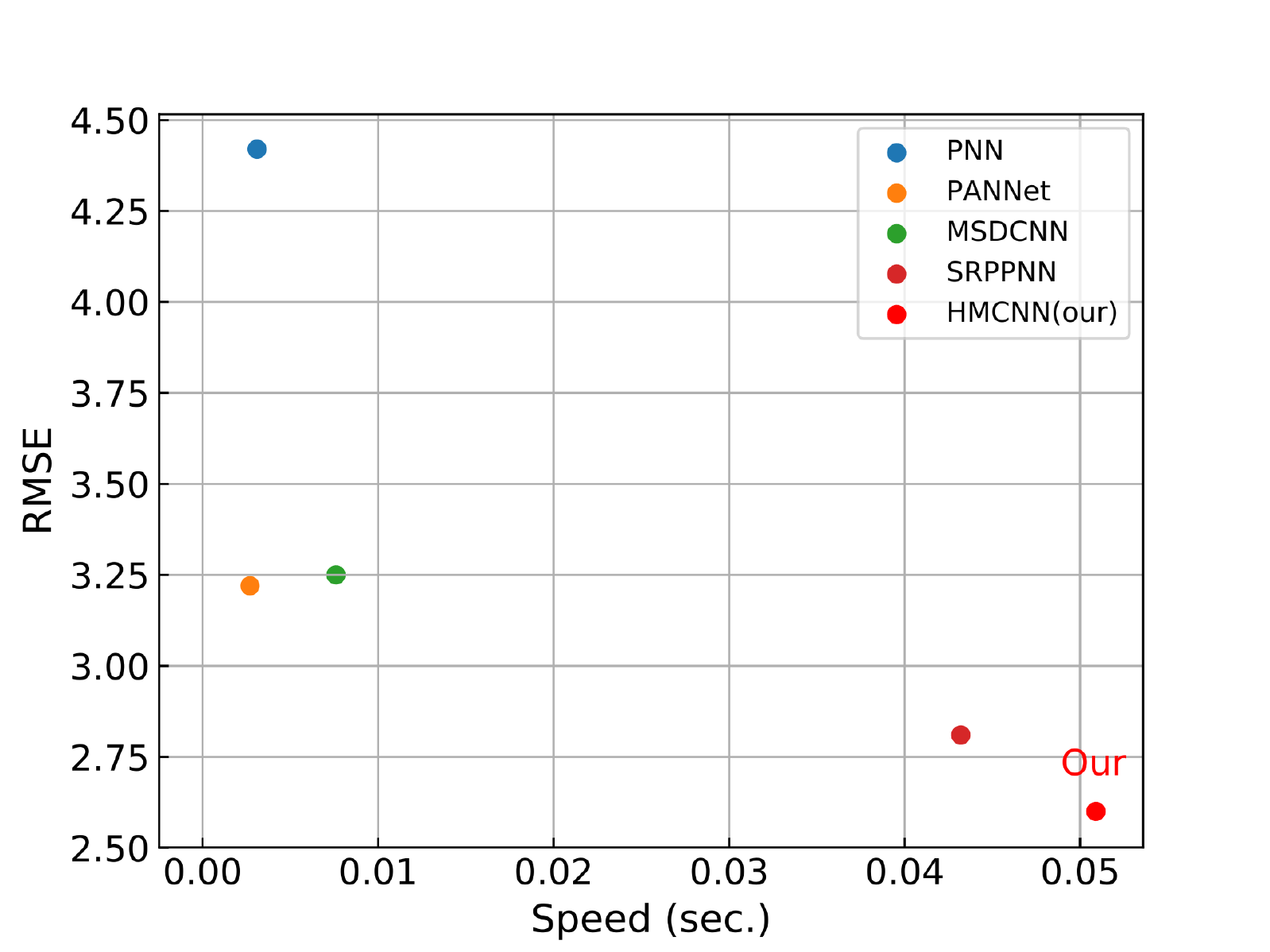}
	\caption{The mean running time for the proposed HMCNN and other deep learning-based methods.}
	\label{time}
\end{figure}

Finally, we investigate the efficiency of HMCNN compared with other deep learning base approaches. The results are shown in Fig. \ref{time}. We notice that SRPPNN and our proposed method require considerably longer processing time than others. For SRPPNN, we believe its dense feature maps and convolution blocks are responsible for the increased complexity. In HMCNN, despite its improved performance, the two high-pass modification blocks demand more computational resources.

\vspace{-0.7cm}
\section{Conclusion}\label{s2}
In this article, we propose a high-pass modification block, termed HMB, for pan-sharpening. We first design the HMB to enhance the high-frequency information in the different bands of the MS images, aiming to restore the rich and detailed information. To further generate visually appealing pan-sharpened images, we introduce a perceptual loss function that targets near-infrared space. The results suggest that the proposed perceptual loss function is able to obtain fine details and sharp edges. Through experiments and ablation studies, the proposed method exhibits state-of-the-art performance.
\vspace{-0.5cm}
\bibliographystyle{IEEEbib}
\bibliography{strings,Template}

\end{document}